\documentclass[11pt]{article}

\usepackage[final]{acl}

\usepackage{times}
\usepackage{latexsym}

\usepackage[T1]{fontenc}

\usepackage[utf8]{inputenc}

\usepackage{microtype}

\usepackage{inconsolata}

\usepackage{graphicx}

\usepackage{amsfonts}
\usepackage{amsmath}
\usepackage{booktabs}
\usepackage{multirow}

\title{MultiToP: Learning to Patch Visual Tokens to Mitigate Hallucinations in Video Large Multimodal Models}

\author{
 \textbf{Yuansheng Gao\textsuperscript{1,$\dagger$,$\ddagger$}},
 \textbf{Wenbin Xing\textsuperscript{2,$\ddagger$}},
 \textbf{Jiahao Yuan\textsuperscript{3}},
 \textbf{Kaiwen Zhou\textsuperscript{1}},
\\
 \textbf{Han Bao\textsuperscript{1,$\S$}},
 \textbf{Zonghui Wang\textsuperscript{1,$\S$}},
 \textbf{Wenzhi Chen\textsuperscript{1}}
\\
\\
 \textsuperscript{1}Zhejiang University,
 \textsuperscript{2}Sun Yat-sen University,
 \textsuperscript{3}East China Normal University
\\
 \small{
   \textsuperscript{$\dagger$}\textbf{Email:} \href{mailto:y.gao@zju.edu.cn}{y.gao@zju.edu.cn}
 }
\\
 \small{
   \textsuperscript{$\ddagger$}Co-first authors
 }
 \small{
   \textsuperscript{$\S$}Corresponding authors
 }
}

\begin{document}
\maketitle
\begin{abstract}
Video Large Multimodal Models have achieved remarkable progress in video understanding, yet they remain prone to hallucinations, where generated responses are not faithfully supported by the input video. In this paper, we propose MultiToP, a multimodal-context-aware visual token patching framework that mitigates hallucinations by refining unreliable visual tokens before language generation. MultiToP introduces a lightweight Visual Token Patcher to predict token-level replacement distributions and selectively substitute unreliable visual tokens with a dynamic global patch token. To train the patcher effectively, we further propose information-guided rank calibration, which uses answer-conditioned frame-level information cues derived from the backbone to guide token replacement. Combined with ground-truth answer supervision and sparsity regularization, MultiToP enables localized visual evidence refinement without modifying the original model. Extensive experiments demonstrate that MultiToP effectively reduces hallucinations on Vript-HAL with negligible inference overhead, improving the F1 scores of Qwen3-VL-4B-Instruct by 50.60\% over the vanilla model. Meanwhile, MultiToP preserves general video understanding ability, yielding an 18.58\% relative accuracy gain on ActivityNet-QA for Video-LLaVA-7B.

\end{abstract}

\section{Introduction}

Video Large Multimodal Models (VideoLMMs) have recently advanced rapidly, showing powerful understanding and reasoning capabilities~\cite{bai2025qwen3, wang2025internvideo2}. However, their practical reliability is still limited by hallucination, where the generated responses appear plausible but are not faithfully supported by the input video~\cite{bai2024hallucination, sahoo2024comprehensive}. This issue is particularly critical for video understanding, as models must not only recognize visual entities but also track actions, relations, and temporal dynamics across frames~\cite{kong2025mhbench, xing2026learning}.

\begin{figure}[t]
  \centering
  \includegraphics[width=\linewidth]{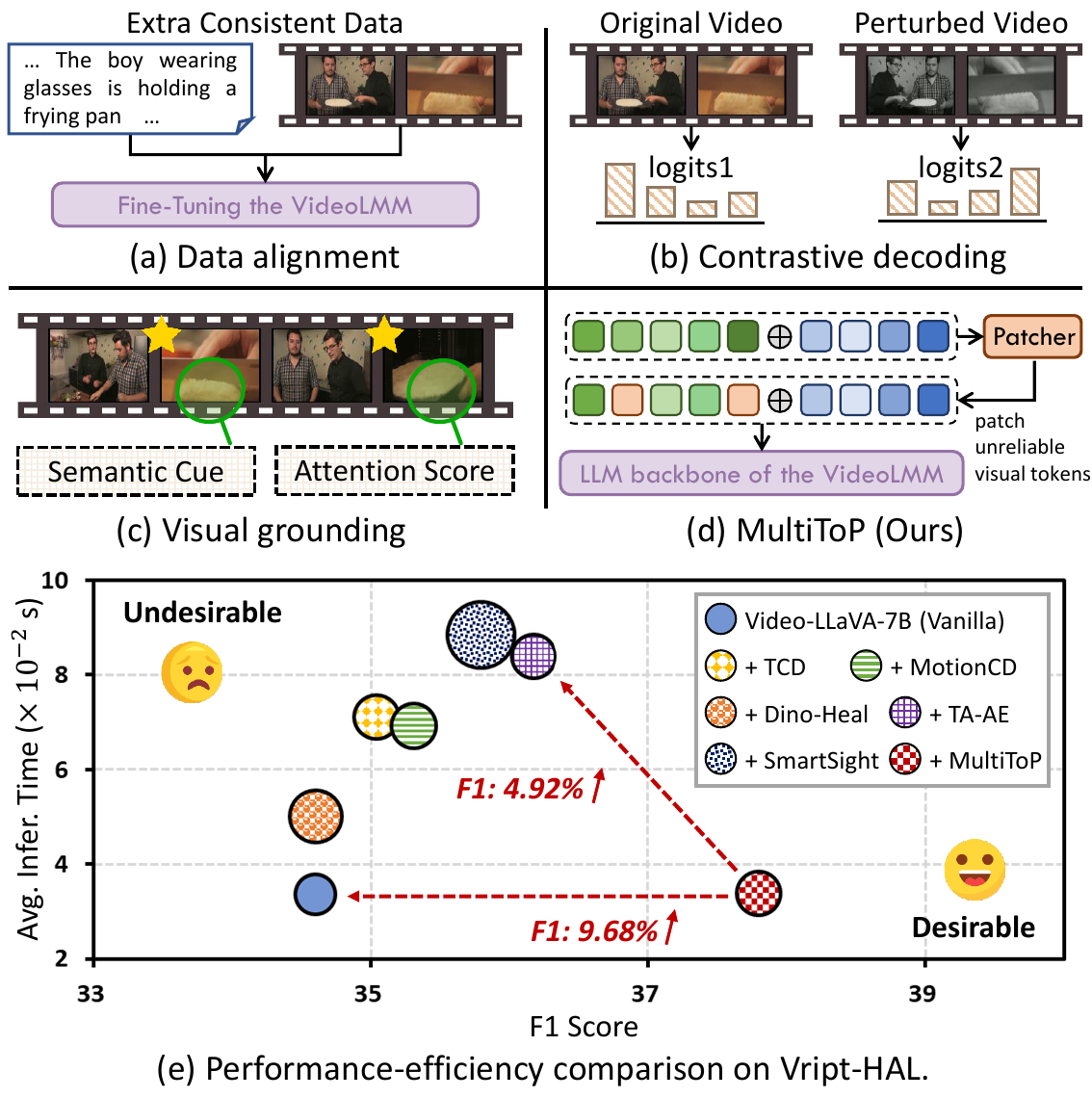}
  \caption{Comparison of different mitigation strategies. Existing methods mainly operate at the data (a), video (b), or frame (c), while MultiToP (d) refines visual evidence at the token-level before generation. For (e), bubble size indicates average GPU memory, and the y-axis represents the normalized inference time per token.}
  \label{fig:motivation}
\end{figure}

Existing studies mitigate hallucinations from different intervention levels, as shown in Figure~\ref{fig:motivation}. Some methods improve video-language alignment through supervised fine-tuning, preference optimization, or reinforcement learning~\cite{bansal2024videocon, gao2025exploring, huang2025vistadpo, ding2025pami, li2026videohallu}. Others perform inference-time intervention by contrasting original and perturbed videos~\cite{leng2024mitigating, zhang2024eventhallusion, kong2025mhbench}, or enhance visual grounding by reweighting salient regions, injecting semantic cues, modifying internal activations, or exploiting attention signals~\cite{li2025vidhalluc, wang2025mitigating, cai2026mitigating, sun2026smartsight, zhang2026verhallu}. Although effective, these methods mainly operate at the model, video, frame, or response level, leaving the fine-grained reliability of individual visual tokens insufficiently explored.

This gap is important because visual tokens are the direct interface through which video evidence is consumed by the language model. In VideoLMMs, video inputs are encoded and projected into visual tokens before being fed into the language model~\cite{zhang2023video, lin2024video}. However, not all visual tokens provide reliable evidence: some capture salient objects, actions, and temporal transitions, while others may encode background regions, redundant information, uncertain features, or imperfect visual-language alignment~\cite{gong2024damro, woo2025don, bai2025mitigating, seo2026epistemic}. Once unreliable tokens enter the language model, they may interact with biased attention patterns or strong language priors and be amplified during autoregressive generation~\cite{leng2024mitigating, huang2024opera, li2025hidden}. Therefore, hallucinations can potentially be mitigated by repairing a small subset of misleading visual tokens before generation.

Motivated by this insight, we propose \textbf{MultiToP}, a \textbf{Multi}modal-context-aware visual \textbf{To}ken \textbf{P}atching framework for hallucination mitigation in VideoLMMs. MultiToP introduces a lightweight Visual Token Patcher that predicts token-level replacement distributions and generates a dynamic global patch token for selective visual token substitution. To train the patcher effectively, we further propose information-guided rank calibration, which uses answer-conditioned frame-level information cues derived from the VideoLMM itself to guide token replacement. Together with ground-truth prediction supervision and sparsity regularization, it enables effective optimization of the patcher. By refining unreliable visual tokens before language generation, MultiToP mitigates hallucinations without modifying the original VideoLMM and introduces negligible inference overhead. Our main contributions are summarized as follows:
\begin{itemize}
    \item We propose MultiToP, a multimodal-context-aware visual token patching framework that mitigates hallucinations in VideoLMMs by refining unreliable visual tokens.

    \item We design a lightweight Visual Token Patcher that predicts token-level replacement distributions and generates a dynamic global patch token for selective visual token substitution.

    \item We propose information-guided rank calibration, a training strategy that uses frame-level information cues to guide token replacement.

   \item Extensive experiments show that MultiToP mitigates hallucinations with negligible inference overhead, improving F1 scores on Vript-HAL by 9.68\% and 50.60\% over vanilla Video-LLaVA-7B and Qwen3-VL-4B-Instruct, respectively, while yielding an 18.58\% relative accuracy gain on ActivityNet-QA for Video-LLaVA-7B.
\end{itemize}

\section{Related Work}
\subsection{Video Large Multimodal Models}
VideoLMMs typically build upon Large Language Models (LLMs)~\cite{zhao2026survey, gao2026boosting, yuan2026kardia} by connecting them with visual or video encoders, thereby extending multimodal understanding to temporally evolving visual sequences~\cite{tang2025video}. Early models established basic video instruction tuning and temporal modeling abilities~\cite{zhang2023video, maaz2024video, lin2024video}, while recent architectures further improve efficiency, long-context modeling, and visual tokenization~\cite{jin2024chat, li2024llama, xu2024pllava}. Despite the strong capabilities of frontier models such as Qwen3-VL~\cite{bai2025qwen3}, hallucination remains a key obstacle to reliable video understanding~\cite{sahoo2024comprehensive, gao2025exploring}.

\subsection{Hallucinations in VideoLMMs}
Hallucinations in VideoLMMs refer to responses that are inconsistent with the visual evidence in the input video~\cite{bai2024hallucination, rawal2025argus}. They are often caused by language priors, biased vision-language correlations, or weak temporal modeling~\cite{bansal2024videocon, zhang2024eventhallusion}. The dynamic spatiotemporal nature of videos further exacerbates this issue, making models particularly vulnerable to compositional errors among entities, actions, and relations~\cite{xing2026learning, yang2026discovering}. To quantify these failures, recent benchmarks have evaluated hallucinations from different perspectives: VideoHallucer focuses on intrinsic and extrinsic errors~\cite{wang2024videohallucer}, EventHallusion studies event-level discrepancies~\cite{zhang2024eventhallusion}, VidHalluc targets temporal inconsistencies~\cite{li2025vidhalluc}, and VideoHallu evaluates counterfactual robustness with synthetic videos~\cite{li2026videohallu}.

\subsection{Hallucination Mitigation}
Existing hallucination mitigation for VideoLMMs broadly falls into three categories. The first improves model alignment via supervised fine-tuning~\cite{bansal2024videocon, gao2025exploring}, preference optimization~\cite{huang2025vistadpo, ding2025pami}, or reinforcement learning~\cite{huang2025taming, li2026videohallu}, aiming to reduce the reliance on language priors and improve video-response consistency. However, these demand costly retraining and annotated data. The second employs inference-time contrastive decoding~\cite{leng2024mitigating, zhang2024eventhallusion, kong2025mhbench} to suppress predictions driven by spurious priors. Yet, predefined perturbations (e.g., sample and reverse) and extra decoding passes make them less adaptive to diverse hallucinations and increase inference cost. The third is to enhance visual grounding. Some methods reweight salient regions~\cite{li2025vidhalluc}, inject explicit semantic cues~\cite{wang2025mitigating}, or modify temporal-aware internal activations~\cite{cai2026mitigating}. Others exploit attention signals more directly. SmartSight detects temporal attention collapse to select less hallucinated candidate responses~\cite{sun2026smartsight}, while VERHallu reallocates frame-level attention around key events~\cite{zhang2026verhallu}. However, these typically depend on external estimators, multiple candidate generations, or coarse frame/response-level interventions, leaving fine-grained token-level visual evidence insufficiently refined.

Concurrently, ViSSRes~\cite{gao2026enhancingvideorepresentationsspatiotemporalsemantic} focuses on representation-level video enhancement by learning a lightweight residual aligner with spatiotemporal and semantic consistency objectives. It differs from MultiToP in intervention granularity, optimization objective, and technical design. In contrast, MultiToP performs token-level intervention before language generation by predicting replacement distributions over visual tokens and selectively patching unreliable tokens. Moreover, MultiToP introduces information-guided rank calibration to align token replacement with answer-conditioned frame-level information cues.

\section{Methodology}

\begin{figure*}
  \centering
  \includegraphics[width=\linewidth]{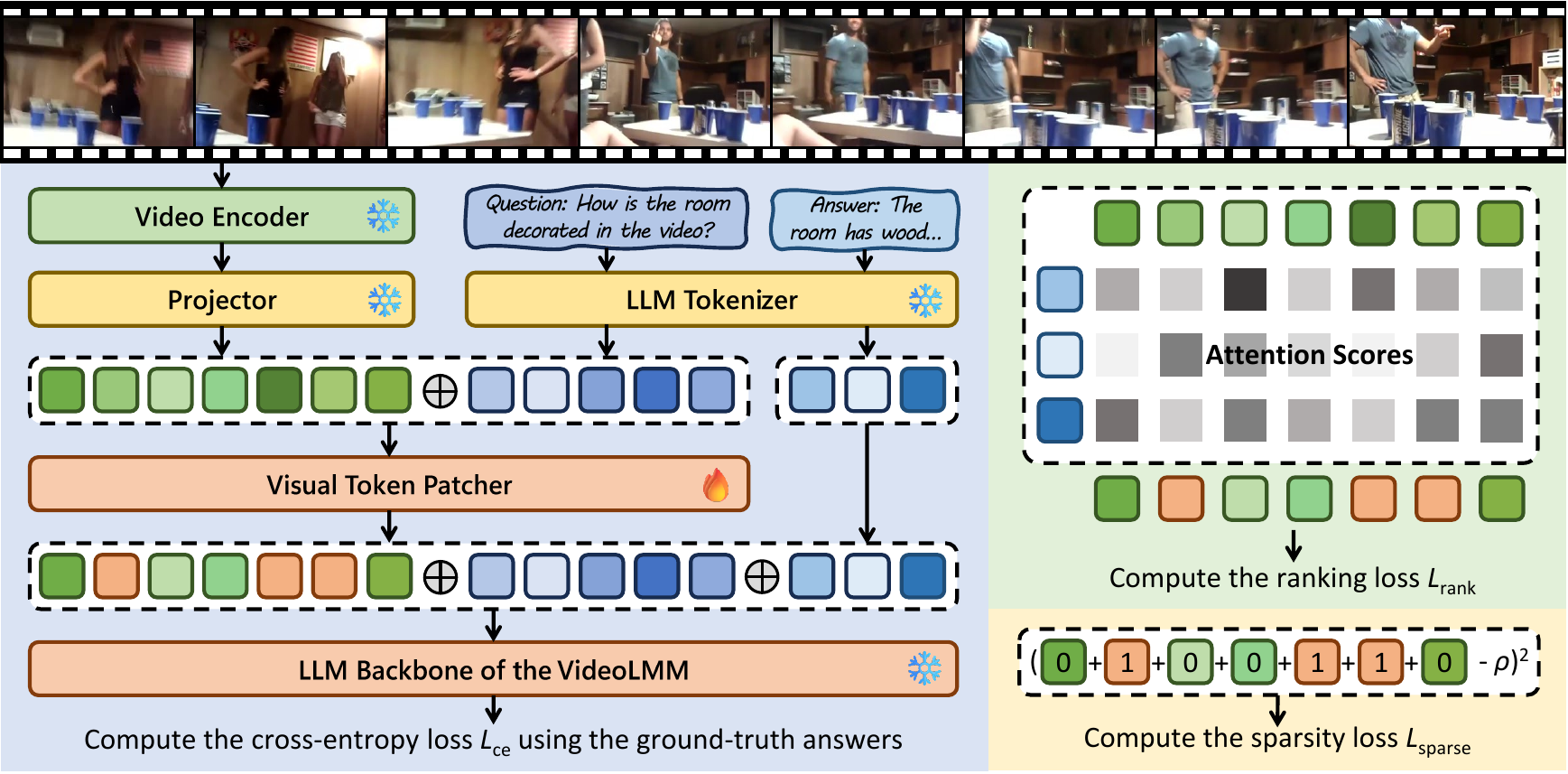}
    \caption{Overview of the proposed MultiToP framework. The original VideoLMM is frozen during training, while only the Visual Token Patcher is optimized. During inference, the trained patcher refines visual tokens before language generation with negligible additional overhead.}
  \label{fig:overview}
\end{figure*}

\subsection{Preliminary}

We consider a VideoLMM parameterized by $\theta$. Given an input video, the model first employs a visual encoder followed by a projector to transform the video into a sequence of visual tokens $v=\{v_m\}_{m=1}^{M} \in \mathbb{R}^{M \times D}$, where $M$ denotes the number of visual tokens and $D$ is the hidden dimension of the VideoLMM. Meanwhile, the user question is tokenized into a sequence of textual tokens $x=\{x_n\}_{n=1}^{N} \in \mathbb{R}^{N \times D}$, where $N$ denotes the number of textual question tokens.

At the $t$-th decoding step, the visual tokens $v$, textual tokens $x$, and previously generated tokens $y_{<t}$ are concatenated and fed into the LLM backbone of the VideoLMM. The next token is predicted in an autoregressive manner, formulated as:
\begin{equation}
\begin{aligned}
    y_t&\sim p_\theta(y_t \mid v, x, y_{<t})\\
    &\propto \exp(\text{logit}_\theta(y_t \mid v,x,y_{<t})),
\end{aligned}
\end{equation}
where $y_t$ denotes the answer token at time step $t$. During the decoding stage, hallucinations often occur when probability mass is incorrectly assigned to tokens that are inconsistent with the visual input.

\subsection{Architecture of the Visual Token Patcher}

MultiToP inserts a lightweight Visual Token Patcher before the autoregressive decoding stage. Given the visual tokens $v$ and textual tokens $x$, we denote the multimodal token sequence as:
\begin{equation}
    z = v \oplus x \in \mathbb{R}^{(M+N)\times D},
\end{equation}
where $\oplus$ denotes concatenation along the token dimension. The patcher takes $z$ as input and produces two outputs: a replacement distribution for identifying unreliable visual tokens, and a dynamic global patch token for token substitution. Although the patcher observes the full multimodal context, it is designed to patch only the visual token span while leaving textual tokens unchanged.

Since visual token reliability depends on both token-level evidence and multimodal context, we use a lightweight Transformer encoder~\cite{vaswani2017attention} to contextualize the input sequence:
\begin{equation}
    h = \mathcal{T}\big(\phi_{\mathrm{down}}(z)\big) \in \mathbb{R}^{(M+N)\times d},
\end{equation}
where $\phi_{\mathrm{down}}$ denotes the token projection layer, $\mathcal{T}$ denotes the lightweight self-attention encoder, $h$ is the contextualized hidden representation of the input sequence, and $d$ is the hidden dimension.

To combine local token evidence with global multimodal information, we split each token representation into two parts. The first half is used as the local representation $h^l$, while the second half is averaged over the whole sequence to obtain a shared global context $h^g$:
\begin{equation}
    h^l_i = h_{i,1:\frac{d}{2}}, 
    \quad
    h^g = \frac{1}{M+N}\sum_{j=1}^{M+N} h_{j,\frac{d}{2}+1:d}.
\end{equation}
The final context-aware representation is then defined as:
\begin{equation}
    \tilde h_i = [h^l_i; h^g].
\end{equation}

Based on $\tilde h_i$, an output MLP predicts a two-dimensional logit vector for each token:
\begin{equation}
    o_i = \phi_{\mathrm{out}}(\tilde h_i) \in \mathbb{R}^{2},
\end{equation}
where the first logit $o_{i,0}$ corresponds to the keep decision and the second logit $o_{i,1}$ corresponds to the replace decision. These logits define the binary replacement distribution used to determine whether each visual token should be preserved or patched.

After obtaining the replacement distribution, we further generate a patch token for visual token substitution. Instead of assigning an individual patch token to each replaced visual token, which would greatly enlarge the optimization space, we construct a dynamic global patch token shared by all replaced visual tokens within the same multimodal context. Specifically, the patch token is generated by combining the averaged visual token representation with a contextual residual:
\begin{equation}
    p = \frac{1}{M}\sum_{m=1}^{M}v_m + \phi_{\mathrm{res}}\Big(\frac{1}{M}\sum_{i\in\mathcal{V}}\tilde h_i\Big) \in \mathbb{R}^{1\times D},
\end{equation}
where $\mathcal{V}$ denotes the positions of visual tokens and $\phi_{\mathrm{res}}(\cdot)$ is a residual projection. The averaged visual representation anchors the patch token in the original space, while the contextual residual enables it to adapt to the current multimodal input.

Overall, the Visual Token Patcher forms a lightweight multimodal-context-aware architecture that jointly predicts token replacement logits and generates a dynamic global patch token, providing the basis for reliable visual token patching in the subsequent training and inference procedure.

\subsection{Visual Token Patching for Refinement}

To achieve efficient visual token patching, we train the patcher to predict whether each visual token should be preserved or replaced. 

\subsubsection{Ground-Truth Answer Supervision}
The main challenge in training the patcher lies in the non-differentiability of applying the $\arg\max$ operation to the replacement distribution. To address this issue, we employ the Gumbel-Softmax~\cite{jang2016categorical} with a temperature parameter $\tau$ to obtain a differentiable binary replacement decision. Specifically, given the replacement logits $o_i \in \mathbb{R}^{2}$ predicted by the patcher, we obtain the replacement indicator $g$ through:
\begin{equation}
    g = \mathrm{GumbelSoftmax}(o, \tau)_1 \in\{0,1\}^{m\times1},
\end{equation}
where the subscript $1$ denotes the replace dimension. In practice, we adopt the straight-through estimator such that $g_i$ behaves as a binary decision in the forward pass while allowing gradients to be back-propagated through the relaxed Gumbel-Softmax distribution. When $g_i = 0$, it indicates that the original visual token is retained, and vice versa. Therefore, the patched visual token sequence $\tilde v$ is then obtained as:
\begin{equation}
    \tilde v = (1-g)\cdot v + g\times p.
\end{equation}
The patched visual tokens are then fed into the original VideoLMM for answer generation.
To directly optimize the replacement distribution and the global patch token for hallucination mitigation, we adopt the standard cross-entropy loss for next-token prediction on the ground-truth answer $y$:
\begin{equation}
    \mathcal{L}_{\mathrm{ce}}
    = - \sum_{t=1}^{L} \log p_\theta(y_t \mid \tilde v, x, y_{<t}),
    \label{eq:loss_ce}
\end{equation}
where $L$ denotes the length of the ground-truth answer sequence.

\subsubsection{Information-Guided Rank Calibration}

Although the cross-entropy loss in Equation~\ref{eq:loss_ce} provides answer-level supervision, it is insufficient to learn an accurate token replacement distribution over multi-frame videos. Since different frames contribute unequally to answering a question, we introduce frame-level information guidance to calibrate the replacement distribution. Intuitively, frames containing less useful information should have a higher tendency to be patched.

Instead of estimating frame relevance with CLIP-like matching~\cite{tang2025adaptive, radford2021learning}, which can be sensitive to explicit object mentions in the question, we derive information cues directly from the VideoLMM. Motivated by~\citet{jiang2025devils, neo2025towards}, we first run the original VideoLMM and collect attention maps from intermediate decoder layers within the normalized depth range of 0.25--0.75. Let $\mathcal{Y}$ denote the positions of the ground-truth answer tokens. We average the attention maps over the selected layers and heads, and extract the attention from answer tokens to visual tokens:
\begin{equation}
    s_i = \frac{1}{|\mathcal{Y}|}\sum_{q \in \mathcal{Y}} A_{q,i}, 
    \quad i \in \mathcal{V},
\end{equation}
where $A_{q,i}$ denotes the aggregated attention weight from answer token position $q$ to visual token position $i$, and $s_i$ is the token-level attention score.

\begin{figure}
  \centering
  \includegraphics[width=0.826\linewidth]{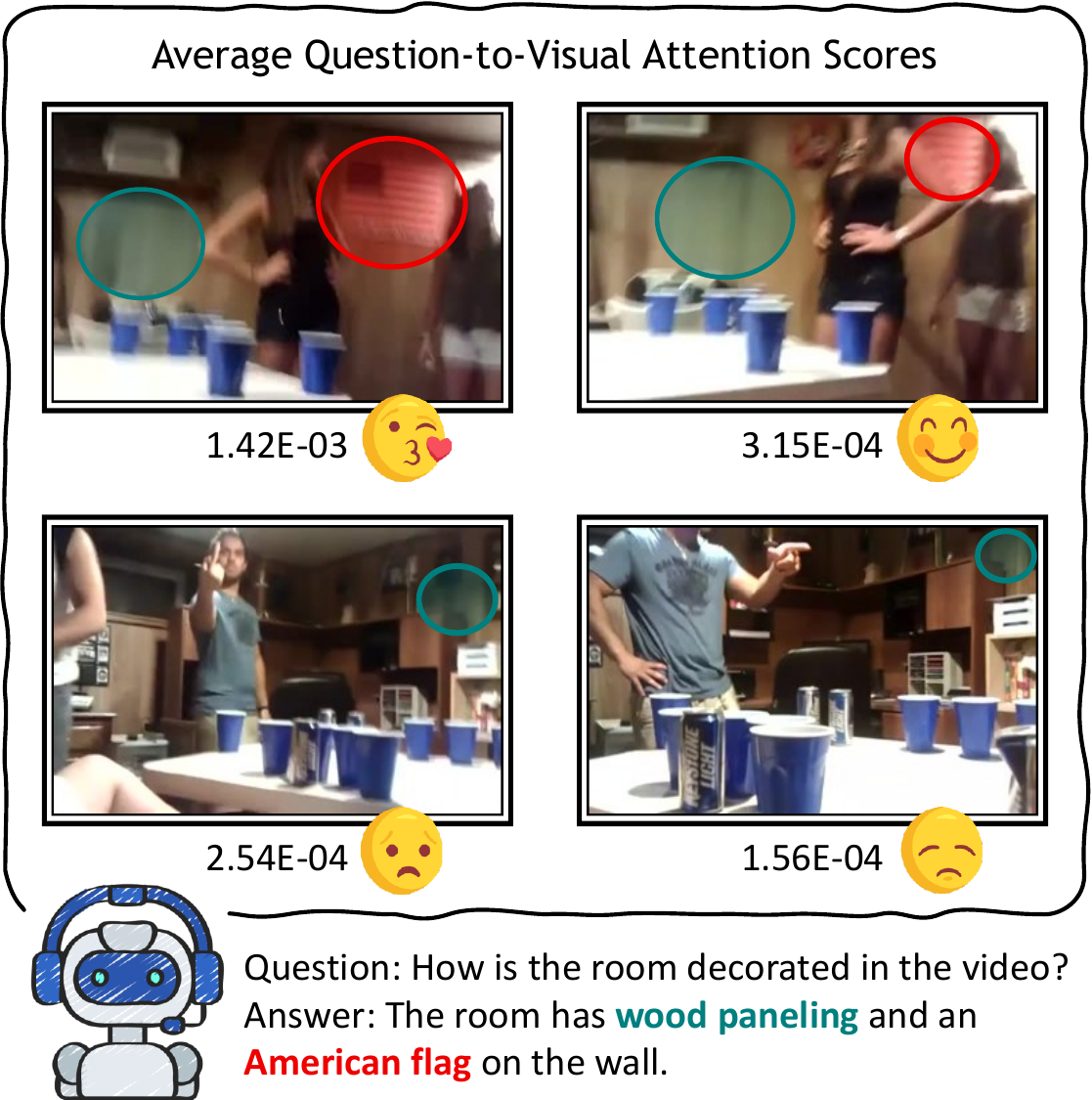}
    \caption{Answer-to-visual attention scores across frames with different levels of useful information. More informative frames receive higher attention (Qwen3-VL-4B-Instruct as the backbone).}
  \label{fig:useful-information}
\end{figure}

After obtaining token-level attention scores, we compute the useful information score for each frame by averaging the top-$K$ most attended visual tokens within that frame. Let $\mathcal{V}_f \subset \mathcal{V}$ denote the visual token positions of the $f$-th frame:
\begin{equation}
    u_f = \frac{1}{K}\sum_{i \in \operatorname{TopK}(\mathcal{V}_f)} s_i,
    \quad f=1,\ldots,F,
    \label{eq:useful_information}
\end{equation}
where $F$ is the number of frames. As shown in Figure~\ref{fig:useful-information}, frames with more useful information tend to receive higher answer-to-visual attention. After that, we normalize the frame-level scores with a softmax function:
\begin{equation}
    a_f = \frac{\exp(u_f)}{\sum_{j=1}^{F}\exp(u_j)},
    \quad
    a = \{a_f\}_{f=1}^{F},
\end{equation}
where $a_f$ is the useful information score of the $f$-th frame. Since these scores are derived from answer-to-visual attention, they are answer-conditioned and model-aware, rather than solely determined by surface-level question-frame similarity.

We next define the frame-level replacement score from the patcher's replacement logits. Given $o_i\in\mathbb{R}^{2}$, we average the replace-over-keep logit margin within each frame:
\begin{equation}
    r_f = \frac{1}{|\mathcal{V}_f|}\sum_{i\in \mathcal{V}_f} \left(o_{i,1} - o_{i,0}\right),
\end{equation}
where a larger $r_f$ indicates a stronger tendency to replace visual tokens in the $f$-th frame.

Rather than matching the absolute values of $a_f$, we use a ranking objective to calibrate the relative replacement tendency across frames. We first define the pairwise ranking weight between frame $f$ and another frame $f'$ as
\begin{equation}
    w_{f,f'} = \operatorname{ReLU}(a_f-a_{f'}),
\end{equation}
which assigns larger weights to frame pairs with larger information gaps. The ranking loss is then formulated as:
\begin{equation}
    \mathcal{L}_{\mathrm{rank}}
    =
    \frac{
    \sum_{f=1}^{F}\sum_{f'=1}^{F}
    w_{f,f'}\operatorname{ReLU}(r_f-r_{f'})
    }{
    \sum_{f=1}^{F}\sum_{f'=1}^{F} w_{f,f'}+\epsilon
    },
\end{equation}
where $\epsilon$ is a small constant for numerical stability. The weight $w_{f,f'}$ emphasizes frame pairs with larger useful-information gaps, which provide more reliable ranking supervision, while down-weighting ambiguous pairs with similar scores. This objective encourages frames with more useful information to obtain lower replacement scores, while the opposite holds for frames with less useful information. Since it only enforces the relative order rather than the exact magnitudes of attention-derived scores, the calibration is less sensitive to noisy attention values and normalization effects.

\subsubsection{Sparsity Regularization}
To control the overall patching budget, we regularize the replacement ratio of visual tokens. Let $\rho$ denote the target replacement ratio. The sparsity loss is defined as:
\begin{equation}
    \mathcal{L}_{\mathrm{sparse}}
    =
    \left(
    \frac{1}{|\mathcal{V}|}
    \sum_{i\in\mathcal{V}} g_i
    -
    \rho
    \right)^2 .
    \label{eq:loss_sparse}
\end{equation}
This regularization prevents excessive patching that may remove useful visual evidence, while encouraging the patcher to replace a sparse subset of unreliable visual tokens.

\subsubsection{Training and Inference}

We combine the above three objectives to train the Visual Token Patcher while keeping the original VideoLMM frozen. The final training objective is formulated as:
\begin{equation}
    \mathcal{L}
    =
    \mathcal{L}_{\mathrm{ce}}
    +\lambda_{\mathrm{rank}}\mathcal{L}_{\mathrm{rank}}
    +\lambda_{\mathrm{sparse}}\mathcal{L}_{\mathrm{sparse}},
    \label{eq:total_loss}
\end{equation}
where $\lambda_{\mathrm{rank}}$ and $\lambda_{\mathrm{sparse}}$ balance the rank calibration loss and the sparsity constraint, respectively.

During inference, the trained patcher directly predicts the replacement logits and generates the global patch token for each test video-question pair. The patched visual tokens are then fed into the frozen VideoLMM together with the textual tokens for autoregressive answer generation. Since no ground-truth answer or teacher attention computation is required, MultiToP only introduces the lightweight overhead of the Visual Token Patcher.

\section{Experiments}
This section presents our key experimental results, covering hallucination mitigation, general capability preservation, ablation studies, computational efficiency, and the impact of cross-dataset training. Additional experiments on robustness across model scales and case studies are provided in Appendices~\ref{sec:add_exp_res} and~\ref{sec:case_study}, respectively.

\begin{table*}
\centering
\small
\begin{tabular}{lcccccccccc}
\toprule
\multirow{2}{*}{Method} 
& \multicolumn{3}{c}{VideoHallucer} 
& \multicolumn{3}{c}{Vript-HAL} 
& \multicolumn{4}{c}{VidHalluc} \\
\cmidrule(lr){2-4} \cmidrule(lr){5-7} \cmidrule(lr){8-11}
& Overall & Pct.~Diff & FP~Ratio 
& Recall & Precision & F1 
& BQA & MCQ & STH & TSH \\
\midrule
Video-LLaVA-7B 
& 16.39 & 0.38 & 0.94 
& 25.45 & 54.00 & 34.60 
& 23.88 & 65.18 & 30.12 & 28.83 \\
+ TCD 
& 17.21 & 0.38 & 0.94 
& 25.82 & 54.49 & 35.04 
& 24.81 & 64.56 & 30.08 & 26.06 \\
+ MotionCD 
& 23.16 & 0.27 & 0.81 
& 25.94 & 55.26 & 35.31 
& 24.74 & 64.00 & 28.60 & 25.92 \\
+ DINO-HEAL 
& 18.14 & 0.37 & 0.94 
& 25.82 & 52.44 & 34.60 
& 24.96 & \textbf{65.76} & 36.57 & 29.24 \\
+ SmartSight 
& 15.00 & 0.35 & 0.97 
& 25.75 & 58.67 & 35.79 
& 23.72 & 65.04 & 37.64 & 30.30 \\
+ TA-AE 
& 14.45 & 0.33 & 0.92 
& 25.94 & 59.74 & 36.17 
& 23.86 & 65.41 & 38.90 & 30.48 \\
+ Ours 
& \textbf{24.08} & \textbf{0.29} & \textbf{0.86} 
& \textbf{26.62} & \textbf{68.28} & \textbf{37.95} 
& \textbf{25.08} & 65.22 & \textbf{42.02} & \textbf{49.33} \\
\midrule
Qwen3-VL-4B 
& 47.69 & -0.09 & 0.34 
& 27.00 & 62.97 & 37.57 
& 58.57 & 86.32 & 37.30 & 56.50 \\
+ TCD 
& 50.47 & -0.11 & 0.30 
& 25.33 & 63.98 & 35.92 
& 59.82 & 85.31 & 37.17 & 55.00 \\
+ MotionCD 
& 42.79 & -0.19 & 0.24 
& 25.51 & 64.42 & 36.17 
& 60.54 & 84.57 & 35.34 & 57.16 \\
+ DINO-HEAL 
& 50.57 & -0.33 & 0.16 
& 26.54 & 63.73 & 37.10 
& 61.08 & 86.90 & 38.10 & 58.30 \\
+ SmartSight 
& 52.70 & -0.08 & 0.36 
& 28.51 & 64.51 & 39.32 
& 59.05 & 85.95 & 40.10 & 58.84 \\
+ TA-AE 
& 49.41 & -0.10 & 0.32 
& 29.11 & 61.77 & 39.09 
& 61.39 & 86.53 & 41.67 & 60.00 \\
+ Ours 
& \textbf{57.58} & \textbf{-0.03} & \textbf{0.44} 
& \textbf{48.21} & \textbf{70.68} & \textbf{56.58} 
& \textbf{74.64} & \textbf{89.08} & \textbf{61.35} & \textbf{86.83} \\
\bottomrule
\end{tabular}
\caption{Comparison results with SOTA inference-time intervention methods. Pct.~Diff and FP~Ratio are better when they are closer to 0 and 0.5, respectively, while higher values are better for all other metrics.}
\label{tab:main_results}
\end{table*}

\subsection{Experimental Settings}
\paragraph{Models and Baselines} We use Video-LLaVA-7B~\cite{lin2024video} and Qwen3-VL-4B-Instruct~\cite{bai2025qwen3} as backbones, and compare MultiToP with a series of state-of-the-art (SOTA) inference-time intervention methods, including TCD~\cite{zhang2024eventhallusion}, MotionCD~\cite{kong2025mhbench}, DINO-HEAL~\cite{li2025vidhalluc}, SmartSight~\cite{sun2026smartsight}, and TA-AE~\cite{cai2026mitigating}. To ensure fairness, we adopt the original settings of these methods.

\paragraph{Evaluation Benchmarks} We use VideoHallucer~\cite{wang2024videohallucer}, Vript-HAL~\cite{yang2024vript}, and VidHalluc~\cite{li2025vidhalluc} to evaluate the hallucination mitigation capability. Meanwhile, Video-MMMU~\cite{hu2025video} and ActivityNet-QA~\cite{yu2019activitynet} are used to assess the general video understanding ability.

\paragraph{Implementation Details} We randomly sample 3,000 samples from LLaVA-Hound dataset~\cite{zhang2024direct} to train the Visual Token Patcher using AdamW~\cite{loshchilov2017decoupled}. The $K$ in Equation~\ref{eq:useful_information} is set to 32. The $\rho$ in Equation~\ref{eq:loss_sparse} is set to 0.3 and 0.2 for Video-LLaVA-7B and Qwen3-VL-4B-Instruct, respectively. The $\lambda_{\mathrm{rank}}$ and $\lambda_{\mathrm{sparse}}$ in Equation~\ref{eq:total_loss} are set to 1 and 100, respectively. Details can be found in Appendix~\ref{sec:implementation}.

\subsection{Performance Evaluation}

We first evaluate the hallucination mitigation ability of MultiToP, and the results are reported in Table~\ref{tab:main_results}. MultiToP consistently improves both vanilla models across the three hallucination benchmarks and achieves the best performance on most metrics. Notably, on Vript-HAL, MultiToP improves the F1 score over the best baseline by 4.92\% with Video-LLaVA-7B and by 43.90\% with Qwen3-VL-4B-Instruct. These results demonstrate the effectiveness of our MultiToP for hallucination mitigation.

We further evaluate whether MultiToP preserves general video understanding ability. As shown in Table~\ref{tab:general_video_understanding}, MultiToP improves both vanilla models on Video-MMMU and ActivityNet-QA. For Video-LLaVA-7B, MultiToP increases the Overall score on Video-MMMU from 13.44 to 14.56 and improves the accuracy on ActivityNet-QA from 40.90 to 48.50. For Qwen3-VL-4B-Instruct, MultiToP also improves the Video-MMMU Overall score from 46.67 to 47.22 and the ActivityNet-QA accuracy from 61.19 to 62.18. These results indicate that MultiToP mitigates hallucinations without sacrificing the general video understanding ability of the original models. The complete metric results are provided in Appendix~\ref{sec:add_exp_res}.

\begin{table}[t]
\centering
\small
\begin{tabular}{lcc}
\toprule
Method & Video-MMMU & ActivityNet-QA \\
\midrule
Video-LLaVA-7B & 13.44 & 40.90 \\
+ Ours & \textbf{14.56} & \textbf{48.50} \\
\midrule
Qwen3-VL-4B & 46.67 & 61.19 \\
+ Ours & \textbf{47.22} & \textbf{62.18} \\
\bottomrule
\end{tabular}
\caption{Comparison between MultiToP and vanilla models on Video-MMMU (Overall) and ActivityNet-QA (Accuracy) benchmarks.}
\label{tab:general_video_understanding}
\end{table}

\subsection{Ablation Study}
In this subsection, we ablate the most critical hyperparameters, including $\rho$ in Equation~\ref{eq:loss_sparse} and $\lambda_\mathrm{rank}$ in Equation~\ref{eq:total_loss}, with Qwen3-VL-4B-Instruct as the backbone. Ablations of Top-$K$ in Equation~\ref{eq:useful_information} and $\lambda_\mathrm{sparse}$ in Equation~\ref{eq:total_loss} are provided in Appendix~\ref{sec:add_exp_res}.

\paragraph{The effect of $\rho$} As shown in Table~\ref{tab:rho_ablation}, $\rho$ has a clear impact on the final performance. Setting $\rho=0$ removes target replacement and leads to the worst result, indicating that this component is necessary. Increasing $\rho$ to 0.1 and 0.2 substantially improves the Overall score and moves Pct.~Diff and FP~Ratio closer to their ideal values. The best performance is achieved at $\rho=0.2$. However, larger values such as 0.3 and 0.4 lead to slight performance degradation, suggesting that excessive target replacement may weaken the useful target signal.

\begin{table}
\centering
\small
\begin{tabular}{lccc}
\toprule
$\rho$ & Overall & Pct.~Diff & FP~Ratio \\
\midrule
0.4   & 55.02 &	-0.04 &	0.42 \\
0.3   & 54.51 &	-0.04 &	0.41  \\
0.2   & \textbf{57.58} &	\textbf{-0.03} &	\textbf{0.44}  \\
0.1   & 57.07 &	\textbf{-0.03} &	0.43  \\
0.0   & 47.69 &	-0.09 &	0.34  \\
\bottomrule
\end{tabular}
\caption{Effect of $\rho$ on VideoHallucer. Pct.~Diff and FP~Ratio are better when they are closer to 0 and 0.5, respectively, while higher values are better for Overall.}
\label{tab:rho_ablation}
\end{table}

\paragraph{The effect of $\lambda_\mathrm{rank}$} As shown in Table~\ref{tab:lambda_rank_ablation}, $\lambda_\mathrm{rank}$ affects the balance between the ranking loss and other training objectives. Without the ranking loss, the Overall score drops to 56.35, showing its positive contribution. The best performance is achieved when $\lambda_\mathrm{rank}=1$, with the highest Overall score of 57.58 and favorable Pct.~Diff and FP~Ratio values. However, larger values such as 5 and 10 lead to lower Overall scores, indicating that an overly strong ranking constraint may disturb the balance of the overall objective.

\begin{table}
\centering
\small
\begin{tabular}{lccc}
\toprule
$\lambda_\mathrm{rank}$ & Overall & Pct.~Diff & FP~Ratio \\
\midrule
0   & 56.35 & -0.03 & 0.44 \\
0.5 & 55.33 & -0.04 & 0.42 \\
1   & \textbf{57.58} & \textbf{-0.03} & \textbf{0.44} \\
2   & 56.97 & \textbf{-0.03} & 0.43 \\
5   & 55.12 & -0.04 & 0.41 \\
10  & 55.53 & \textbf{-0.03} & 0.43 \\
\bottomrule
\end{tabular}
\caption{Effect of $\lambda_\mathrm{rank}$ on VideoHallucer. Pct.~Diff and FP~Ratio are better when they are closer to 0 and 0.5, respectively, while higher values are better for Overall.}
\label{tab:lambda_rank_ablation}
\end{table}

\subsection{Computational Efficiency Analysis}
To evaluate the computational overhead of MultiToP, we report the efficiency comparison in Table~\ref{tab:efficiency_comparison}. Our method introduces only negligible latency compared with the vanilla models. On Video-LLaVA-7B, MultiToP achieves a similar average inference time to the vanilla baseline ($3.39 \times 10^{-2}$ vs. $3.37 \times 10^{-2}$ s/token), while being faster than all other enhanced variants. On Qwen3-VL-4B-Instruct, it also maintains near-baseline latency and is faster than TCD, MotionCD, DINO-HEAL, SmartSight, and TA-AE. Although MultiToP slightly increases memory usage over the vanilla models, its memory footprint remains comparable to lightweight baselines such as TCD, MotionCD, and TA-AE, and is much lower than DINO-HEAL and SmartSight. These results show that MultiToP improves robustness with minimal computational overhead.

\begin{table}[t]
\centering
\small
\begin{tabular}{lcc}
\toprule
Method & Ave. Time/s & Memory/GB \\
\midrule
Video-LLaVA-7B & $3.37 \times 10^{-2}$ & 15.74 \\
+ TCD          & $7.11 \times 10^{-2}$ & 17.32 \\
+ MotionCD     & $6.93 \times 10^{-2}$ & 17.09 \\
+ DINO-HEAL    & $5.01 \times 10^{-2}$ & 20.62 \\
+ SmartSight   & $8.85 \times 10^{-2}$ & 28.55 \\
+ TA-AE        & $8.40 \times 10^{-2}$ & 16.82 \\
+ Ours         & $3.39 \times 10^{-2}$ & 16.52 \\
\midrule
Qwen3-VL-4B          & $3.79 \times 10^{-2}$ & 8.38 \\
+ TCD                & $8.78 \times 10^{-2}$ & 9.14 \\
+ MotionCD           & $9.15 \times 10^{-2}$ & 9.15 \\
+ DINO-HEAL          & $4.71 \times 10^{-2}$ & 10.67 \\
+ SmartSight         & $5.34 \times 10^{-2}$ & 13.54 \\
+ TA-AE              & $3.99 \times 10^{-2}$ & 8.97 \\
+ Ours               & $3.82 \times 10^{-2}$ & 9.10 \\
\bottomrule
\end{tabular}
\caption{Efficiency comparison on the Vript-HAL dataset. The average inference time is computed over all generated tokens, and the memory usage is averaged from samples collected every 2 seconds.}
\label{tab:efficiency_comparison}
\end{table}

\subsection{Cross-Dataset Training Analysis}
To examine whether MultiToP depends on a specific training dataset, we train the patcher on different datasets with the same data scale and training settings. As shown in Table~\ref{tab:cross_dataset_res}, MultiToP consistently improves the vanilla model across all training datasets, suggesting that its effectiveness is not tied to a particular corpus. On Vript-HAL, all three training datasets substantially improve the F1 score, with ShareGPT4Video~\cite{chen2024sharegpt4video} achieving the best result and LLaVA-Video-178k~\cite{zhang2024llava} also bringing a clear gain over the vanilla model. On ActivityNet-QA, all variants improve accuracy, where LLaVA-Hound performs best, while ShareGPT4Video and LLaVA-Video-178k still maintain better performance than the vanilla baseline. These results indicate that MultiToP is not tied to a specific training corpus, though dataset choice affects its balance between hallucination mitigation and general video understanding. The complete results are provided in Appendix~\ref{sec:add_exp_res}.

\begin{table}
\centering
\small
\begin{tabular}{lcc}
\toprule
Training Dataset & Vript-HAL & ActivityNet-QA \\
\midrule
Vanilla & 34.60 &40.90 \\
LLaVA-Hound & 37.95 &\textbf{48.50}  \\
ShareGPT4Video & \textbf{42.22} &46.08 \\
LLaVA-Video-178k & 36.39 &44.44  \\
\bottomrule
\end{tabular}
\caption{Cross-dataset training analysis on Vript-HAL (F1 score) and ActivityNet-QA (Accuracy) under the same training scale and settings.}
\label{tab:cross_dataset_res}
\end{table}

\section{Conclusion}
In this paper, we propose MultiToP, a multimodal-context-aware visual token patching framework for hallucination mitigation in VideoLMMs. MultiToP uses a lightweight Visual Token Patcher to refine unreliable visual tokens before language generation. We further propose information-guided rank calibration to guide token replacement using answer-conditioned frame-level information cues derived from the VideoLMM itself. Experiments on multiple benchmarks demonstrate that MultiToP consistently outperforms existing inference-time intervention methods in hallucination mitigation while preserving the models' general video understanding ability. In terms of efficiency, MultiToP introduces only negligible additional inference overhead, achieving a favorable performance-efficiency trade-off. These findings highlight the effectiveness of token-level visual evidence refinement for building more reliable VideoLMMs.

\section*{Limitations}
Although MultiToP effectively mitigates hallucinations in VideoLMMs, it relies on a predefined target replacement ratio in the sparsity regularization to control the patching budget. However, the optimal replacement ratio may vary across videos and questions, depending on factors such as visual complexity, temporal dynamics, and visual-textual alignment reliability. As a result, using a fixed global replacement ratio may lead to insufficient patching for samples with many unreliable visual tokens, or excessive replacement for samples where most visual evidence is already reliable. This suggests that an adaptive patching budget may be beneficial for further improving MultiToP.


\bibliography{custom}

\appendix

\section{Implementation Details}
\label{sec:implementation}

\paragraph{Additional Experimental Setup} All experiments were conducted using 8 NVIDIA RTX A6000 48GB GPUs. During training, we use a batch size of 1 with 4-step gradient accumulation. The patcher is trained for 2 epochs with a learning rate of $1\times10^{-5}$. To stabilize optimization, we further apply gradient clipping to the predictor parameters with a maximum norm of 1.0. When using Video-LLaVA-7B as the backbone, we set the number of frames for the visual token patcher to 8 during both training and inference. When using Qwen3-VL-4B-Instruct as the backbone, due to resource constraints, we set the number of frames to 16 during training, while following the default setting during inference. When evaluating on Vript-HAL, we set the maximum number of tokens to 1024, while for the other datasets, we set it to 512. For datasets and metrics that require LLMs as evaluators, we adopt the evaluator models and parameter settings specified in their original papers. For all evaluations, we follow the standard evaluation protocols of the corresponding original papers.

\paragraph{Adaptation to Qwen3-VL}
Qwen3-VL adopts the DeepStack mechanism~\cite{meng2024deepstack}, where intermediate visual features are injected into early language layers. To keep the patched visual tokens consistent with these DeepStack features, we adapt MultiToP to perform replacement in the high-resolution visual token space. Specifically, the patcher still predicts replacement logits and a dynamic patch token from the multimodal input embeddings. The replacement logits are predicted over the LLM-side low-resolution visual token positions, while the generated patch token is further mapped to the high-resolution visual token dimension by an additional projection layer:
\begin{equation}
    p^{\mathrm{hr}} = \phi_{\mathrm{hr}}(p).
\end{equation}
The low-resolution keep mask is then expanded according to the spatial merge unit and applied to the corresponding high-resolution visual tokens before spatial merging. The patched high-resolution tokens are subsequently processed by the original Qwen3-VL vision encoder, producing both the final visual embeddings and the DeepStack visual features. Thus, MultiToP can be applied to Qwen3-VL with only a lightweight high-resolution projection layer added.

\section{Additional Experimental Results}
\label{sec:add_exp_res}

\subsection{Additional Ablation Study}
\paragraph{The effect of Top-$K$}
As shown in Table~\ref{tab:topk_ablation}, Top-$K$ affects the balance between informative candidate selection and noise introduction. The best performance is obtained when Top-$K=32$, achieving the highest Recall, Precision, and F1 scores. Smaller values may provide insufficient guidance, while a larger value such as 64 introduces more noisy candidates and leads to a noticeable performance drop.

\begin{table}
\centering
\small
\begin{tabular}{lccc}
\toprule
Top-$K$ & Recall & Precision & F1 \\
\midrule
16   & 46.61 &	70.64 &	55.26 \\
24   & 46.92 &	70.01 &	55.28 \\
32   & \textbf{48.21} &	\textbf{70.68} &	\textbf{56.58} \\
48   & 46.39 &	70.37 &	55.00 \\
64   & 44.19 &	69.38 &	53.09 \\
\bottomrule
\end{tabular}
\caption{Effect of Top-$K$ on Vript-HAL. Higher values are better for all metrics.}
\label{tab:topk_ablation}
\end{table}

\paragraph{The effect of $\lambda_\mathrm{sparse}$}
As shown in Table~\ref{tab:lambda_sparse_ablation}, $\lambda_\mathrm{sparse}$ affects the balance between sparse regularization and the main optimization objective. Without the sparse loss, the Overall score is only 53.89, showing its positive contribution. The best performance is obtained when $\lambda_\mathrm{sparse}=100$, achieving the highest Overall score of 57.58 and the best Pct.~Diff and FP~Ratio values. Larger weights such as 200 and 500 lead to lower performance, indicating that excessive sparse regularization may weaken other useful optimization signals.

\begin{table}
\centering
\small
\begin{tabular}{lccc}
\toprule
$\lambda_\mathrm{sparse}$ & Overall & Pct.~Diff & FP~Ratio \\
\midrule
0    & 53.89 & -0.05 & 0.41 \\
10   & 51.02 & -0.05 & 0.40 \\
50   & 55.84 & -0.04 & 0.41 \\
100  & \textbf{57.58} & \textbf{-0.03} & \textbf{0.44} \\
200  & 55.43 & -0.04 & 0.42 \\
500  & 56.05 & -0.04 & 0.41 \\
\bottomrule
\end{tabular}
\caption{Effect of $\lambda_\mathrm{sparse}$ on VideoHallucer. Pct.~Diff and FP~Ratio are better when they are closer to 0 and 0.5, respectively, while higher values are better for Overall.}
\label{tab:lambda_sparse_ablation}
\end{table}

\subsection{Complete Results}
The complete results for Table~\ref{tab:cross_dataset_res} are reported in Table~\ref{tab:cross_dataset_complete}. The complete results for Table~\ref{tab:general_video_understanding} are reported in Table~\ref{tab:general_video_understanding_complete}.

\begin{table*}
\centering
\small
\begin{tabular}{lccccc}
\toprule
\multirow{2}{*}{Dataset} 
& \multicolumn{3}{c}{Vript-HAL} 
& \multicolumn{2}{c}{ActivityNet-QA} \\
\cmidrule(lr){2-4} \cmidrule(lr){5-6}
& Recall & Precision & F1 & Accuracy & Score \\
\midrule
Vanilla & 25.45 & 54.00 & 34.60 & 40.90 & 2.95 \\
LLaVA-Hound & 26.62 & 68.28 & 37.95 & 48.50 & 3.15 \\
ShareGPT4Video & 31.33 & 67.14 & 42.22 & 46.08 & 3.11 \\
LLaVA-Video-178k & 25.30 & 67.90 & 36.39 & 44.44 & 2.98 \\
\bottomrule
\end{tabular}
\caption{Complete results of cross-dataset training analysis on Vript-HAL and ActivityNet-QA under the same training scale and settings.}
\label{tab:cross_dataset_complete}
\end{table*}

\begin{table*}
\centering
\small
\begin{tabular}{lcccccc}
\toprule
\multirow{2}{*}{Method} 
& \multicolumn{4}{c}{Video-MMMU} 
& \multicolumn{2}{c}{ActivityNet-QA} \\
\cmidrule(lr){2-5} \cmidrule(lr){6-7}
& Perception & Comprehension & Adaptation & Overall 
& Accuracy & Score \\
\midrule
Video-LLaVA-7B & 10.00 & \textbf{11.67} & 20.00 & 13.44 & 40.90 & 2.95 \\
+ Ours      & \textbf{12.00} & 11.33 & \textbf{20.33} & \textbf{14.56} & \textbf{48.50} & \textbf{3.15} \\
Qwen3-VL-4B    & 62.00 & 47.00 & \textbf{31.00} & 46.67 & 61.19 & 3.36 \\
+ Ours      & \textbf{64.67} & \textbf{48.00} & 29.00 & \textbf{47.22} & \textbf{62.18} & \textbf{3.56} \\
\bottomrule
\end{tabular}
\caption{Complete results of general video understanding evaluation on Video-MMMU and ActivityNet-QA.}
\label{tab:general_video_understanding_complete}
\end{table*}

\subsection{Robustness across Model Scales}
To evaluate the robustness of MultiToP across different model scales, we conduct experiments on VidHalluc using Qwen3-VL-2B-Instruct, Qwen3-VL-4B-Instruct, and Qwen3-VL-8B-Instruct. As shown in Table~\ref{tab:size_comparison}, MultiToP consistently improves performance across all model sizes. Specifically, it improves the average score across four metrics by 14.71, 18.30, and 9.51 points on the 2B, 4B, and 8B models, respectively. The gains are particularly pronounced on STH, where MultiToP improves the vanilla models by 33.43, 24.05, and 22.71 points across the three scales. These results demonstrate that MultiToP is not tied to a specific model size and can robustly mitigate hallucinations for both smaller and larger VideoLMMs.

\begin{table}
\centering
\small
\begin{tabular}{llcccc}
\toprule
Size & Method & BQA & MCQ & STH & TSH \\
\midrule
\multirow{2}{*}{2B} 
& Vanilla & 57.86 & 80.89 & 25.00 & 50.00 \\
& Ours    & 75.98 & 85.52 & 58.43 & 52.67 \\
\midrule
\multirow{2}{*}{4B} 
& Vanilla & 58.57 & 86.32 & 37.30 & 56.50 \\
& Ours    & 74.64 & 89.08 & 61.35 & 86.83 \\
\midrule
\multirow{2}{*}{8B} 
& Vanilla & 64.42 & 87.25 & 37.29 & 65.00 \\
& Ours    & 76.35 & 88.96 & 60.00 & 66.67 \\
\bottomrule
\end{tabular}
\caption{Performance comparison across different Qwen3-VL model sizes on VidHalluc.}
\label{tab:size_comparison}
\end{table}

\section{Case Study}
\label{sec:case_study}

To provide an intuitive understanding of how MultiToP mitigates hallucinations, we present representative qualitative examples from multiple datasets.

\begin{figure}
  \centering
  \includegraphics[width=0.9\linewidth]{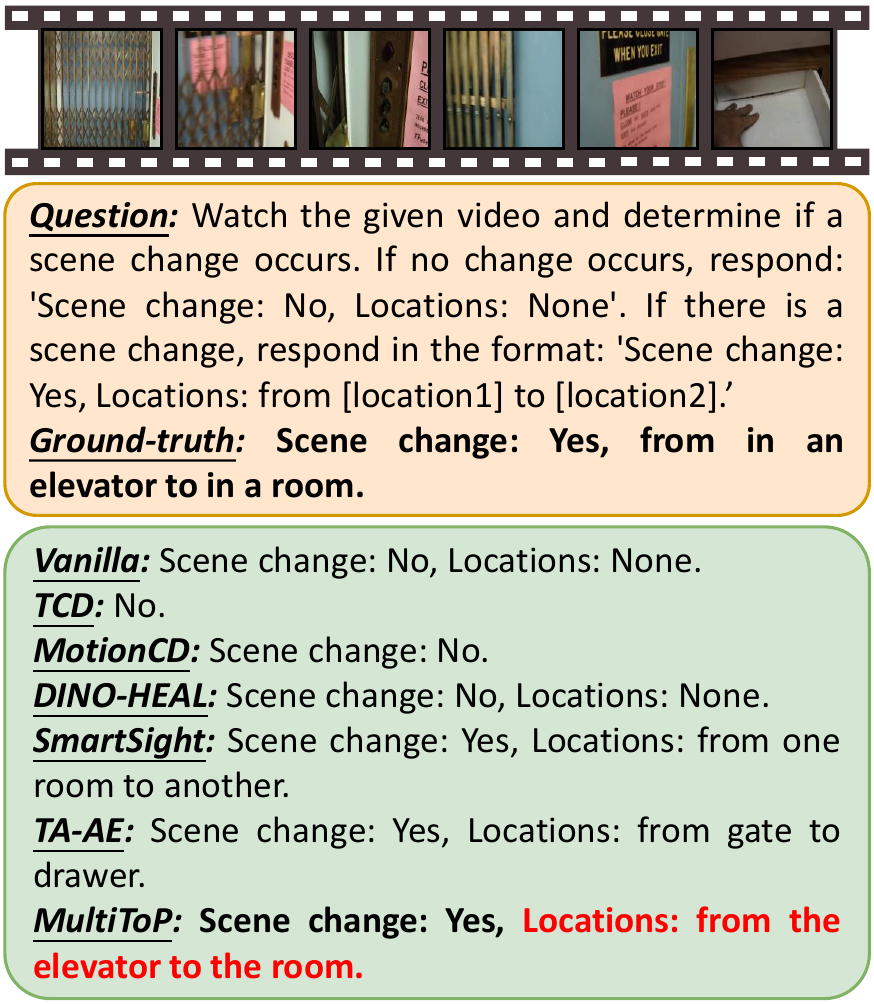}
  \caption{A representative case from the VidHalluc dataset using Qwen3-VL-4B-Instruct as the backbone.}
  \label{fig:case_2_1}
\end{figure}

As shown in Figure~\ref{fig:case_2_1}, the model is tasked with determining whether a scene change occurs in the video and identifying the corresponding locations. The ground-truth clearly indicates a transition from an elevator to a room. However, the Vanilla model, alongside several baseline methods including TCD, MotionCD, and DINO-HEAL, completely fails to detect this temporal transition, incorrectly predicting that no scene change occurred. While SmartSight and TA-AE successfully recognize the existence of a scene change, they suffer from severe object-level hallucinations regarding the specific locations: SmartSight generates a vague description ("from one room to another"), and TA-AE outputs entirely incorrect semantics ("from gate to drawer"). In stark contrast, MultiToP accurately captures both the temporal transition and the precise spatial locations, correctly outputting "from the elevator to the room".

\begin{figure}
  \centering
  \includegraphics[width=0.9\linewidth]{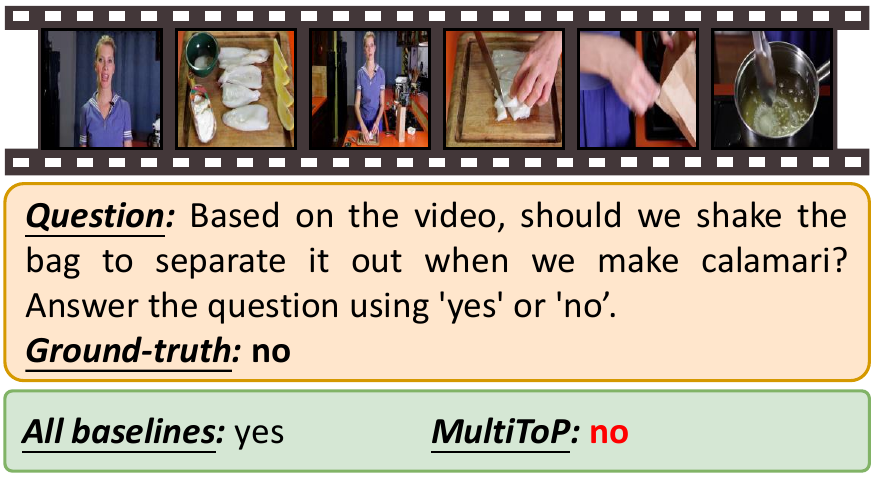}
  \caption{A representative case from the VideoHallucer dataset using Qwen3-VL-4B-Instruct as the backbone.}
  \label{fig:case_1_1}
\end{figure}

\begin{figure}
  \centering
  \includegraphics[width=0.9\linewidth]{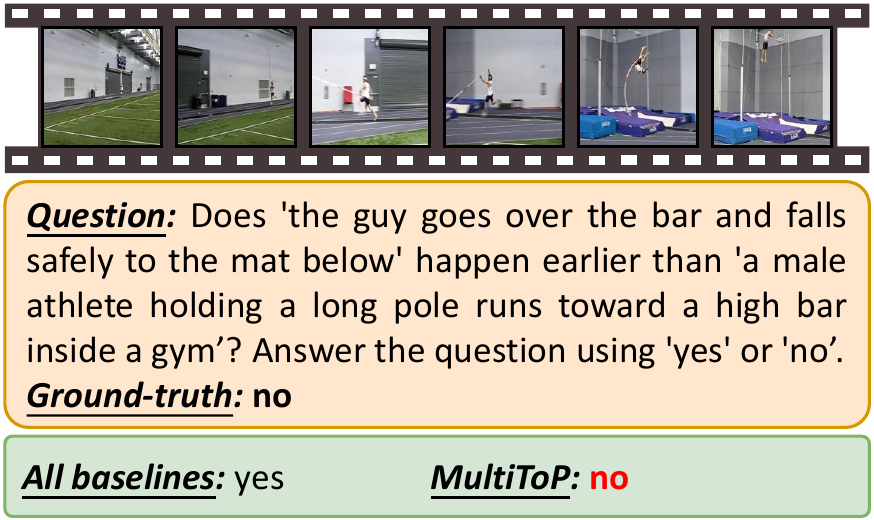}
  \caption{A representative case from the VideoHallucer dataset using Video-LLaVA-7B as the backbone.}
  \label{fig:case_1_2}
\end{figure}

\begin{figure}
  \centering
  \includegraphics[width=0.9\linewidth]{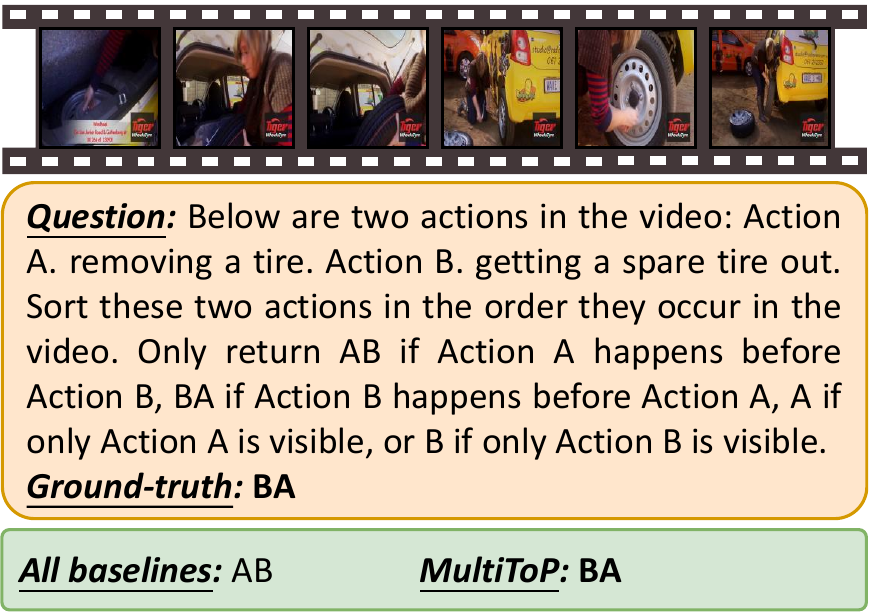}
  \caption{A representative case from the VidHalluc dataset using Video-LLaVA-7B as the backbone.}
  \label{fig:case_2_2}
\end{figure}

Other representative cases are shown in Figures~\ref{fig:case_1_1},~\ref{fig:case_1_2},~\ref{fig:case_2_2},~\ref{fig:case_3_1}, and~\ref{fig:case_3_2}. These qualitative results further demonstrate the effectiveness of MultiToP.

\newpage

\begin{figure*}
  \centering
  \includegraphics[width=0.95\linewidth]{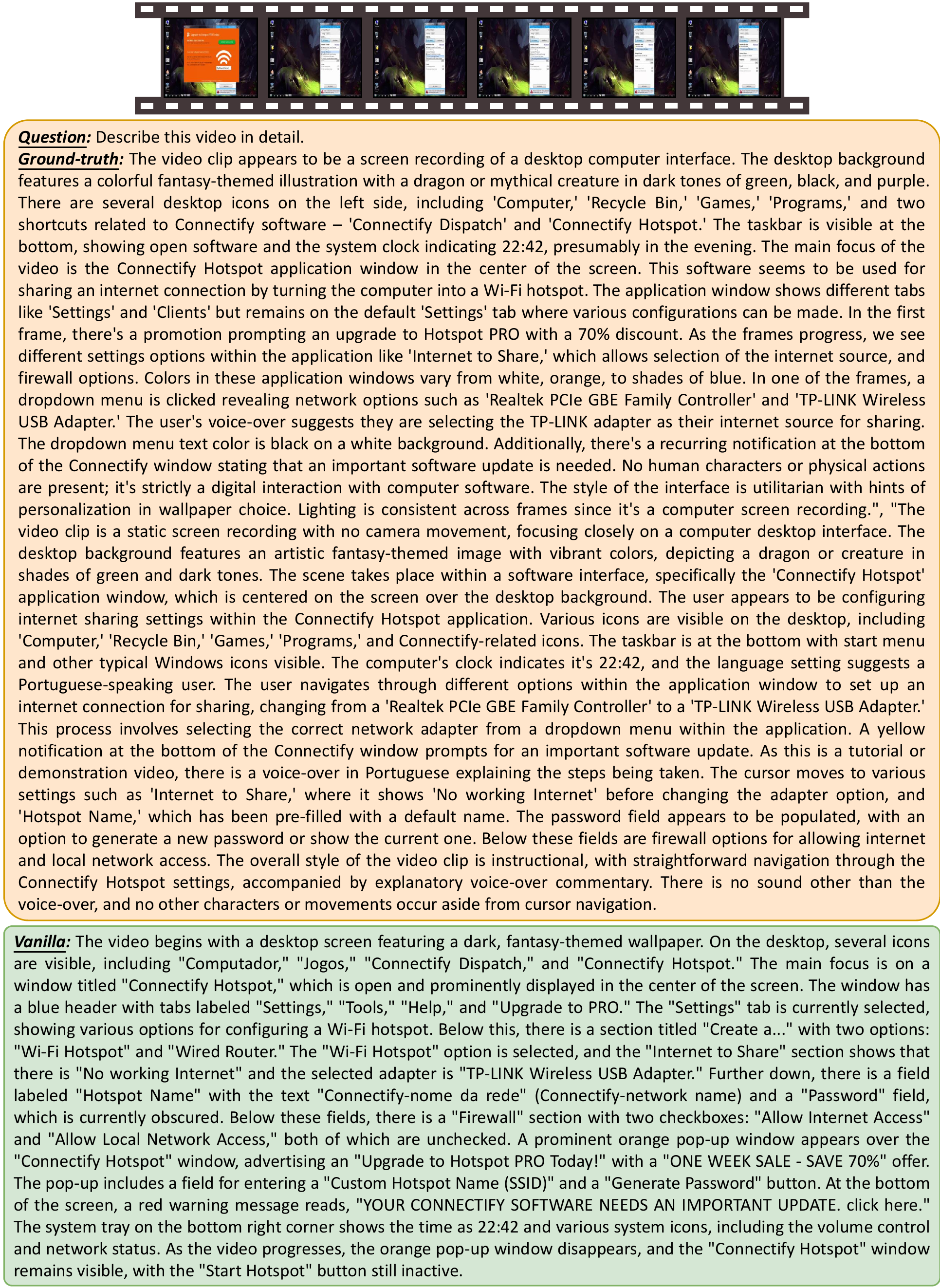}
  \caption{A representative case from the Vript-HAL dataset using Qwen3-VL-4B-Instruct as the backbone.}
  \label{fig:case_3_1}
\end{figure*}

\newpage

\begin{figure*}
\ContinuedFloat
  \centering
  \includegraphics[width=0.95\linewidth]{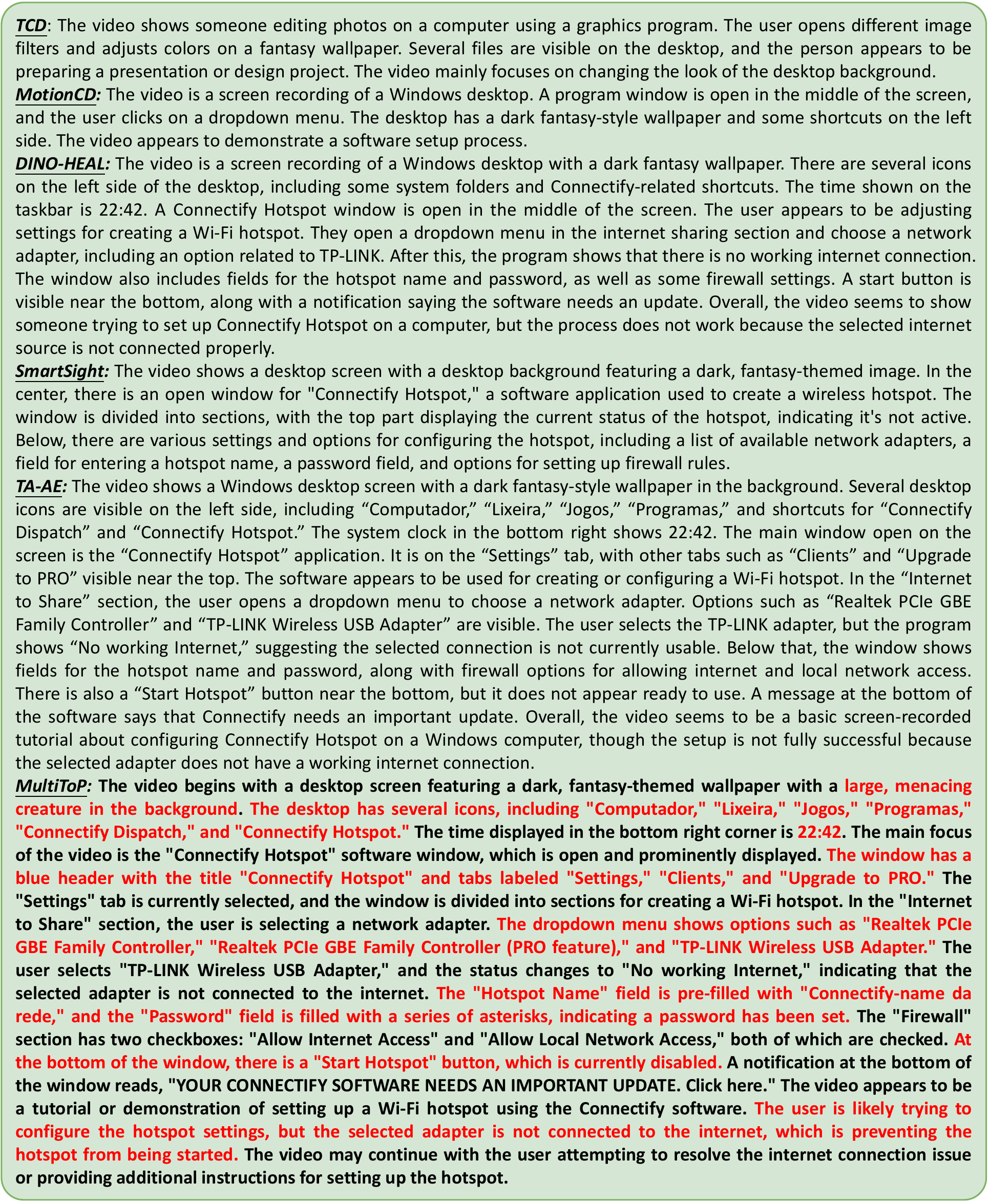}
  \caption{A representative case from the Vript-HAL dataset using Qwen3-VL-4B-Instruct as the backbone.}
  \label{fig:case_3_1}
\end{figure*}

\newpage

\begin{figure*}
  \centering
  \includegraphics[width=0.95\linewidth]{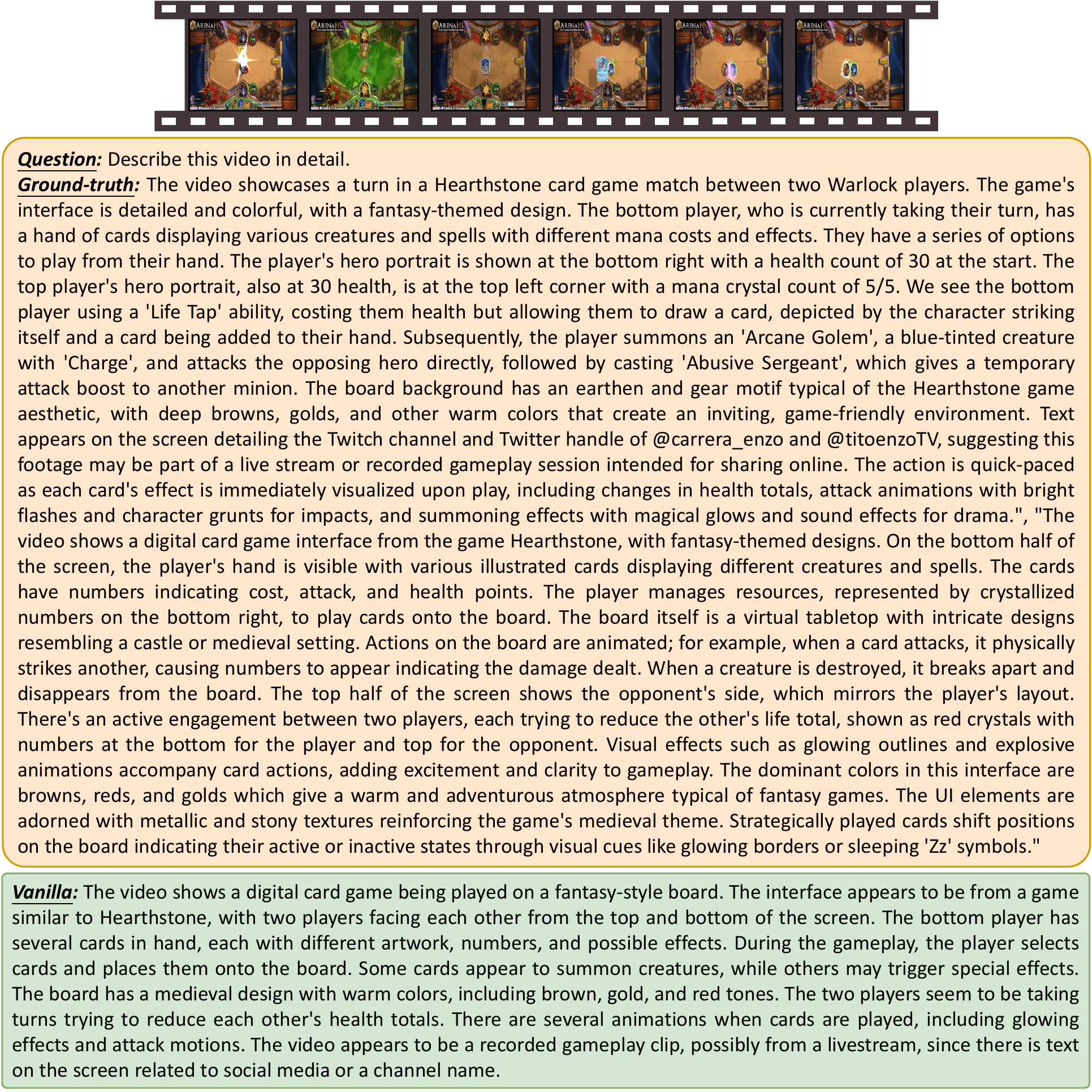}
  \caption{A representative case from the Vript-HAL dataset using Video-LLaVA-7B as the backbone.}
  \label{fig:case_3_2}
\end{figure*}

\newpage

\begin{figure*}
  \ContinuedFloat
  \centering
  \includegraphics[width=0.95\linewidth]{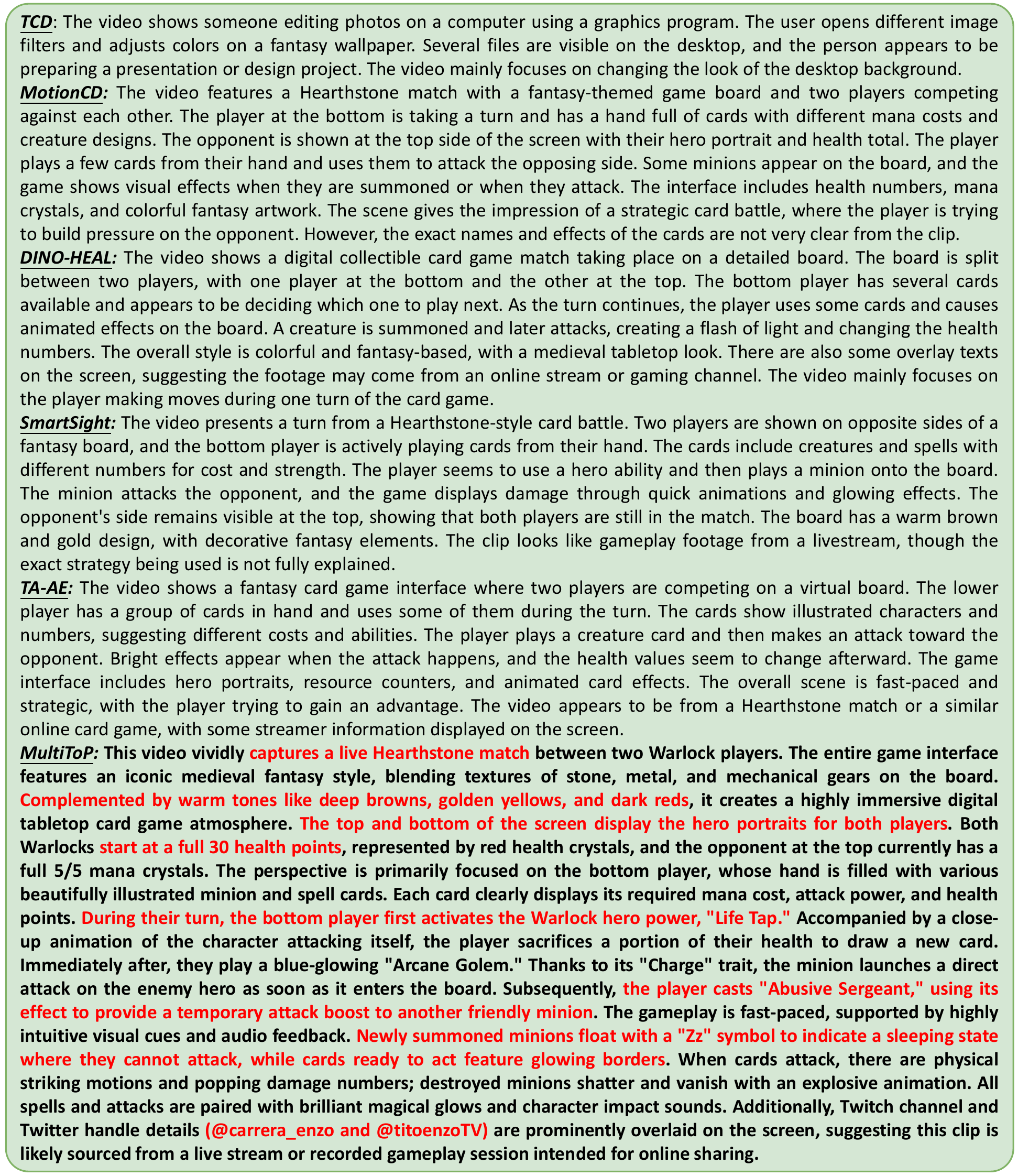}
  \caption{A representative case from the Vript-HAL dataset using Video-LLaVA-7B as the backbone.}
  \label{fig:case_3_2}
\end{figure*}

\end{document}